\documentclass[runningheads]{llncs}
\usepackage[T1]{fontenc}
\usepackage{graphicx}

\usepackage{marvosym}

\usepackage{cuted}
\usepackage{amsmath}
\usepackage{hyperref}
\usepackage{pifont}
\usepackage{marvosym}
\usepackage{amssymb}
\usepackage{subcaption}
\usepackage{booktabs}
\usepackage{multirow}
\usepackage{graphicx}
\usepackage{xcolor}
\usepackage{placeins}
\usepackage[table]{xcolor}

\begin{document}
\title{IMAC-AgriVLN: Can Agricultural Vision-and-Language Navigation Agents be Aware of Instruction Mistakes?}
\titlerunning{IMAC-AgriVLN}
%

\author{
Xiaobei Zhao\textsuperscript{\rm 1,2}\textsuperscript{(\Letter)} \and
Xingqi Lyu\textsuperscript{\rm 1} \and
Xin Chen\textsuperscript{\rm 1,2} \and
Xiang Li\textsuperscript{\rm 1,2}
}

\authorrunning{Xiaobei Zhao et al.}

\institute{\textsuperscript{\rm 1}China Agricultural University\\
\textsuperscript{\rm 2}China Agricultural University-Sichuan Advanced Agricultural \& Industrial Institute\\
\email{\{xiaobeizhao2002,lxq99725\}@163.com, \{chxin,cqlixiang\}@cau.edu.cn}}




\maketitle              

\begin{figure}[h]
\vspace{-5pt}
\centering
\includegraphics[width=1.0\columnwidth]{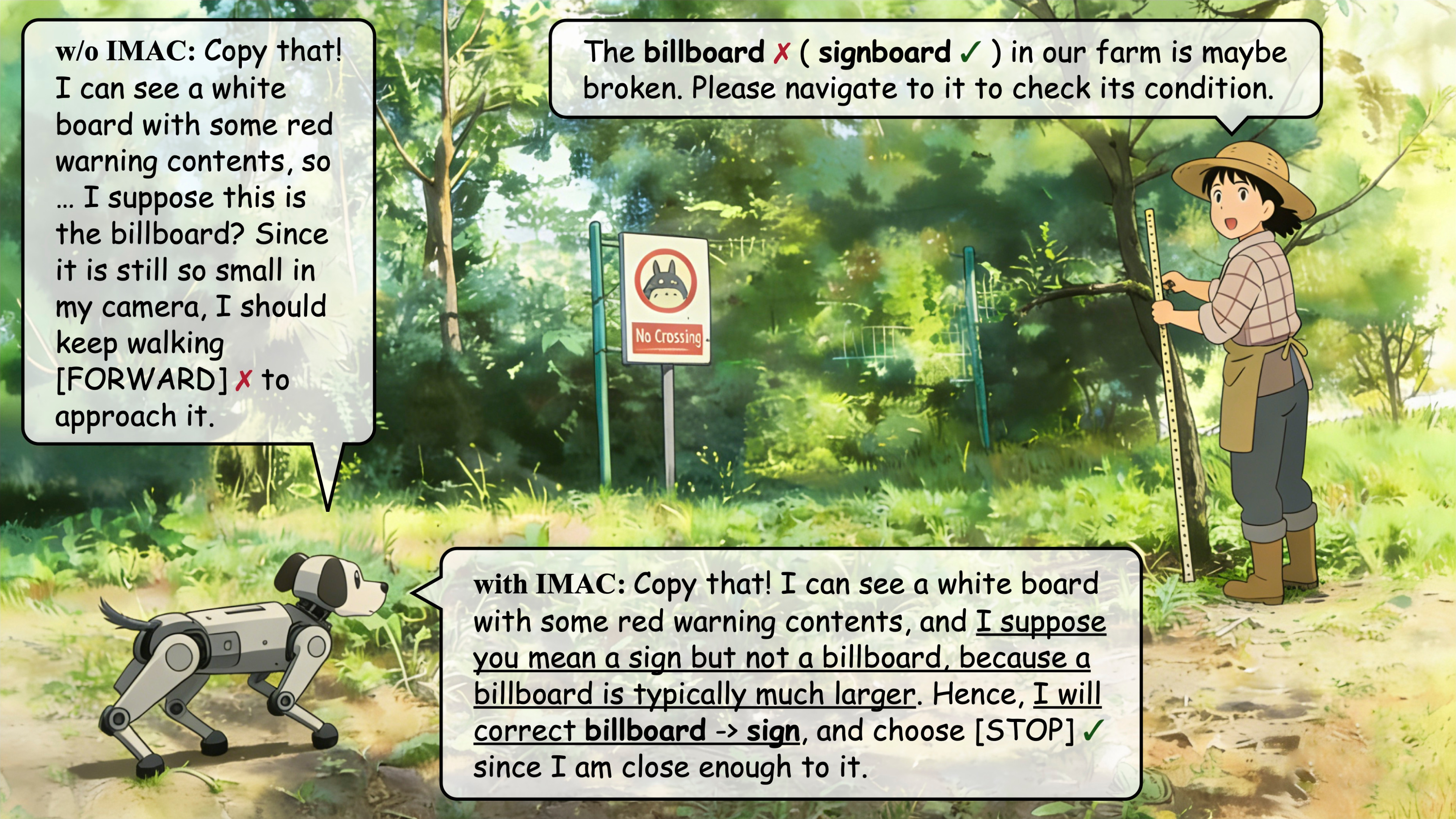} 
\caption{Our A2A-MI benchmark and our IMAC module illustration: When the instruction is inserted with a mistake, the decision-maker cannot succeed this navigation anymore. After IMAC is integrated, it can be aware of the instruction mistake and correct billboard \textcolor[HTML]{B11E31}{\ding{55}} $\rightarrow$ sign \textcolor[HTML]{135416}{\ding{51}} in a reasonable manner, thereby assisting the decision-maker to succeed this navigation again.}
\label{fig:teaser}
\vspace{-20pt}
\end{figure}

\begin{abstract}
Agricultural robots are playing as important roles across a wide range of tasks, nevertheless, they are still mainly depending on manual operations or fixed railways for moving. The A2A benchmark and the AgriVLN method pioneeringly extended Vision-and-Language Navigation (VLN) to the agricultural domain, successfully navigating agricultural robots from starting points to target positions following natural language instructions. However, we observed that almost all the prior VLN methods adopted an ideal assumption: The given instructions themselves were correct. This assumption did not align with the realistic scenarios, because anybody might say an instruction with mistakes, which raised us a question: If an instruction had a mistake, could an agricultural VLN agent be aware of it? To answer this question, we propose the A2A-MI benchmark, in which we follow A2A as the foundation benchmark and insert three classes of instruction mistakes. We use it to evaluate several state-of-the-art agricultural VLN agents, then observe sufficient drops across all of them, such as AgriVLN decreases SR by 57\% in average and increases NE by 9\% in average, from which we suggest the lacking awareness on instruction mistakes. To address this problem, we propose the IMAC module analyzing the instruction and image, to reason whether the instruction has mistakes and attempt to correct them when needed. We integrate it into the AgriVLN backbone to build our IMAC-AgriVLN method, successfully saving SR from 0.10 to 0.14 and NE from 4.81 m to 4.79 m, which demonstrates the effectiveness of IMAC on strengthening the robustness against instruction mistakes. 
Project: \href{https://github.com/AlexTraveling/IMAC-AgriVLN}{https://github.com/AlexTraveling/IMAC-AgriVLN}.

\end{abstract}

\section{Introduction}

\par Agricultural robots are playing as important roles across a wide range of tasks, such as pest detection \cite{IJCNN:PestDetection}, phenotypic measurement \cite{ICIC:TomatoScanner}, and fruit harvesting \cite{ICRA:AHPPEBot}. However, they are still mainly depending on manual operations or fixed railways for moving, which results in lacking movability and insufficient adaptability. 

\par In contrast, Vision-and-Language Navigation (VLN) \cite{CVPR:R2R,ECCV:VLN-CE} enables navigating robots from starting points to target positions following natural language instructions, having shown excellent performances across various domains, such as R2R \cite{CVPR:R2R} for daily residences, TOUCHDOWN \cite{CVPR:TOUCHDOWN} for city streets, and AerialVLN \cite{ICCV:AerialVLN} for aerial spaces. Inspired by them, Zhao et al. \cite{arXiv:AgriVLN} built the A2A benchmark and the AgriVLN method to pioneeringly extend VLN to the agricultural domain, thereby effectively alleviating the problems on lacking movability and insufficient adaptability. 

\par However, we observed that almost all the prior VLN methods 
\cite{arXiv:AgriVLN,ICIC:T-araVLN,arXiv:SUM-AgriVLN,ICIC:MDE-AgriVLN,arXiv:TEA-AgriVLN}
adopted an ideal assumption: The given instructions themselves were correct. This assumption did not align with the realistic scenarios, because anybody might say an instruction with mistakes \cite{arXiv:DDTime,ICCV:Frequency}, as illustrated in Fig. \ref{fig:teaser}. If the decision-maker were a human, it had the subconsciousness to doubt whether the instruction had mistakes when the scenes it saw did not align with the instruction it received \cite{Authorea,TEC,arXiv:Allocation}. 
This deviation raised us a question: \textbf{\textit{If an instruction had a mistake, could an agricultural VLN agent be aware of it?}}

\par The only research having studied on this question, as far as we know, was R2RIE-CE \cite{IROS:Error}, which followed VLN-CE \cite{ECCV:VLN-CE} as the foundation benchmark, then artificially injected three types of errors into the instructions, pioneeringly building a benchmark to evaluate the robustness against instruction errors, while we noticed two serious weaknesses: 
1) In terms of data, all the 11,764 episodes relied on artificial annotation, which was inefficient and costly to be reproduced on other benchmarks; 
2) In terms of methods, to detect and localize the instruction errors before starting running, they trained a cross-modal Transformer taking the camera images across all the time steps as inputs. In fact, however, an agent only accessed the initial single camera image at that moment. Hence, we suggested a large gap from simulation to reality in R2RIE-CE \cite{IROS:Error}. 

\par To extend this evaluation for agricultural VLN agents and avoid the current weaknesses of R2RIE-CE \cite{IROS:Error}, we propose the benchmark of \textbf{A}griculture-\textbf{to}-\textbf{A}griculture with \textbf{M}istaken \textbf{I}nstructions (\textbf{A2A-MI}), in which we follow A2A as the foundation benchmark, then build a semi-automatic data annotator to insert three classes of instruction mistakes into each original instruction in a more efficient and diversified way.

\par We use A2A-MI to evaluate several state-of-the-art agricultural VLN agents, then observe sufficient drops across all of them, such as AgriVLN decreases Success Rate (SR) by 57\% in average and increases Navigation Error (NE) by 9\% in average, from which we suggest an answer to the opening question: \textbf{\textit{Agricultural VLN agents tend to assume that the given instructions are correct, so when the scenes they see do not align with the instructions they receive, they do not have the awareness to doubt the instructions, unless the instruction mistakes are extremely obvious.}}

\par To build the awareness on instruction mistakes for them, we propose the module of \textbf{I}nstruction \textbf{M}istake \textbf{A}wareness and \textbf{C}orrection (\textbf{IMAC}), in which we design two core principles to prompt a Vision-Language Model (VLM) analyzing the instruction and the front-facing camera image at the current time step, to reason whether the instruction has mistakes and attempt to correct them when needed. We integrate IMAC into the AgriVLN backbone to build the method of \textbf{Agri}cultural \textbf{V}ision-and-\textbf{L}anguage \textbf{N}avigation with \textbf{I}nstruction \textbf{M}istake \textbf{A}wareness and \textbf{C}orrection (\textbf{IMAC-AgriVLN}), successfully saving SR from 0.10 to 0.14 and NE from 4.81 m to 4.79 m, which demonstrates the effectiveness of IMAC on strengthening the robustness against instruction mistakes. 

\par In summary, our main contributions are as follows: 
\begin{itemize}
\item We propose the A2A-MI benchmark, in which we follow A2A as the foundation benchmark, then build a semi-automatic data annotator to insert three classes of instruction mistakes into each original instruction. 
\item We evaluate several state-of-the-art agricultural VLN agents on A2A-MI, then observe sufficient drops across all of them, from which we suggest the lacking awareness on instruction mistakes. 
\item We propose the IMAC module and integrate it into the AgriVLN backbone to build our IMAC-AgriVLN method, effectively strengthening the robustness against instruction mistakes.
\end{itemize}

\section{Related Works}

\subsection{A2A}
\par Agriculture-to-Agriculture (A2A) \cite{arXiv:AgriVLN} was previously the only one VLN benchmark specially designed for agricultural robots, consisting of 1,560 episodes across 6 types of scenes: farm, greenhouse, forest, mountain, garden and village, in which the instructions belonged to the step-by-step format and the action space belonged to the continuous environment. 

\subsection{AgriVLN}
\par Vision-and-Language Navigation for Agricultural Robots (AgriVLN) \cite{arXiv:AgriVLN} was the first agricultural VLN method, which used NavGPT \cite{AAAI:NavGPT} as the backbone and integrated the Large Language Model (LLM)-based Subtask List (STL) module, enabling navigating agricultural robots from starting points to target positions following natural language instructions.

\section{Benchmark}

In this section, we present the A2A-MI benchmark. First, we introduce the task definition in Sec. \ref{sec:task_definition}. Second, we define the instruction mistake insertion policy in Sec. \ref{sec:policy}. Third, we present the semi-automatic data annotation in Sec. \ref{sec:data_annotation}. Fourth, we assess the proposed benchmark in Sec. \ref{benchmark_assessment}. 

\subsection{Task Definition}
\label{sec:task_definition}

The task of Agricultural VLN \cite{arXiv:AgriVLN} is defined as follows: In each episode, the agent is given an instruction in natural language, denoted as $W = \langle w_1, w_2, w_3, \dots, w_L \rangle$, where $L$ is the number of words. At each time step $t$, the agent is given a front-facing RGB image $I_t$. The purpose is understanding both $W$ and $I_t$, to select the most appropriate low-level action $\hat{a_t}$ from the action space $\{ \texttt{[FORWARD]}$, $\texttt{[LEFT ROTATE]}$, $\texttt{[RIGHT ROTATE]}$, $\texttt{[STOP]} \}$, thereby navigating the agricultural robot from the starting point to the target position. 

\subsection{Instruction Mistake Insertion Policy}
\label{sec:policy}

We follow all the instructions and images of the A2A \cite{arXiv:AgriVLN} benchmark, then remain all the images unchanged and only implement revisions on the instructions. We categorize human's common speaking mistakes into three classes: descriptive adjectives, concrete nouns, and behavioral verbs, basically satisfying all the common scenarios in agricultural VLN. To insert a class of instruction mistake into an episode, we deliberately use a mistaken contiguous span to replace the original one with the same length. An ideal revised instruction should contain both the core semantic and the misleading mistake, i.e., a human can be aware of the exist of the mistake, and can still succeed this navigation.

\par Here we share an example to better explain what an instruction mistake class refers to and how to insert it.
\begin{quote}
Instruction: \textit{There is a warning \underline{signboard} with a \underline{white} background on the mountain. It has a red circle, a black figure and some characters. Please go forward to check its condition. Please rotate to find and face it first, which is on your right. Don't stop rotating until it is in your horizontal center. Then \underline{go} towards it and stop when you are very close to it.}
\end{quote}
\begin{itemize}
\item \textbf{Descriptive Adjective} refers to an adjective span to describe an object, such as a color or a shape. A reasonable option is: \textit{white} \textcolor[HTML]{135416}{\ding{51}} $\rightarrow$ \textit{green} \textcolor[HTML]{B11E31}{\ding{55}}.
\item \textbf{Concrete Noun} refers to a concrete but not abstract object span, such as a plant or an animal. A reasonable option is: \textit{signboard} \textcolor[HTML]{135416}{\ding{51}} $\rightarrow$ \textit{billboard} \textcolor[HTML]{B11E31}{\ding{55}}.
\item \textbf{Behavioral Verb} refers to a specific low-level action span, such as going forward or turning left. A reasonable option is: \textit{go} \textcolor[HTML]{135416}{\ding{51}} $\rightarrow$ \textit{revolve} \textcolor[HTML]{B11E31}{\ding{55}}.
\end{itemize}

\subsection{Semi-Automatic Data Annotation}
\label{sec:data_annotation}
To implement the above instruction mistake insertion policy in a more efficient and diversified way, we present a semi-automatic data annotator, which runs in a two-stage manner for each episode. 

\par In the first stage, we package the policy to a system prompt template denoted as $P_{sys}^{\mathcal{A}}$, in which we design the core content as follows: 
\begin{quote}
\textit{You are an expert in Vision-and-Language Navigation (VLN) for guiding an agricultural robot. 
Now, your task is to deliberately modify a small part of an original instruction, to respectively introduce three classes of mistakes:\\ 
1. descriptive adjective, denoted as MIa.\\
2. concrete noun, denoted as MIn.\\
3. behavioral verb, denoted as MIv.\\
To introduce each class of mistake, you should give three answers. Usually, in one answer, you should try to modify only one word. Only when necessary, you can modify more than one word.
}
\end{quote}
We prompt an LLM denoted as $\mathcal{A}\,(\,\cdot\,)$ in a zero-shot manner, to automatically generate a total of three revision candidates for each instruction mistake class, defined as:
\begin{equation}
\left\{ C_m^k \right\}_{m \in \{adj, nun, vrb\},\, k=1}^3 
= 
\mathcal{A}\big(P_{sys}^{\mathcal{A}}, P_{usr}^{\mathcal{A}} (W)\big)
\end{equation}
where $P_{usr}^{\mathcal{A}}\,(\,\cdot\,)$ is the user prompt template for $\mathcal{A}\,(\,\cdot\,)$. Every $C_m^k$ consists of three parts, defined as:
\begin{equation}
C_m^k = \left(S_m^k,\;\tilde{S}_m^k,\;W_m^{\prime k}\right)
\end{equation}
where $S_m^k$ is a contiguous span of length $\ell$ in $W$, and $\tilde{S}_m^k$ is its revised counterpart with the same length. $W_m^{\prime k}$ is the revised instruction.

\par In the second stage, for each instruction mistake class, we invite a human specialist denoted as $\mathcal{H}\,(\,\cdot\,)$ to inspect all the three candidates, then manually select the most reasonable one, defined as:
\begin{equation}
C_m^{*} = \mathcal{H}\!\left(\left\{ C_m^k \right\}_{k=1}^3\right)
\end{equation}
where $C_m^{*}$ is the candidate selected by the human specialist. 

\par Finally, we package the revised instruction $W_m^{\prime *}$ with the corresponding image sequence $\langle I_0, I_1, I_2, \dots, I_t \rangle$ to form an A2A-MI episode.

\begin{table*}[t]
\captionsetup{skip=8pt}
\caption{Assessment results between our A2A-MI benchmark with several mainstream VLN benchmarks: “Gran.”, “Mis.”, “Len.” and “V.” represent granularity, mistake, length and view, respectively. “I.”, “O.”, “G.”, “A.”, “T.”, “E.”, “P.” and “S.” represent indoor, outdoor, goal-directed, action-directed, training, evaluation, panoramic and single, respectively. 
}
\label{tab:benchmark_summary}
\centering
\resizebox{\linewidth}{!}{
\renewcommand{\arraystretch}{1.3}
\setlength{\tabcolsep}{3pt}
\begin{tabular}{lcc|cccc|ccc}
\toprule
\multirow{2}{*}{\textbf{Benchmark}} &
\multicolumn{2}{c}{\textbf{Scene}} & \multicolumn{4}{c}{\textbf{Instruction}} & \multicolumn{2}{c}{\textbf{Camera}} \\
\cmidrule(lr){2-3} \cmidrule(lr){4-7} \cmidrule(lr){8-9} 
& \textbf{Scope} & \textbf{Domain} & \textbf{Gran.} & \textbf{Mis.} & \textbf{Len.} & \textbf{Number} & \textbf{V.} & \textbf{Image} \\
\midrule
R2R                & I.     & Residence   & A.       & \ding{55} & 29 &  7,189 (T.\&E.) & P. & RGB-D \\
TOUCHDOWN          & O.     & Street      & A.       & \ding{55} & 90 &  9,326 (T.\&E.) & P. & RGB \\
REVERIE            & I.     & Residence   & G.\&A.   & \ding{55} & 18 & 21,702 (T.\&E.) & P. & RGB-D \\
VLN-CE             & I.     & Residence   & A.       & \ding{55} & 19 &  4,475 (T.\&E.) & S. & RGB-D \\
AerialVLN          & O.     & Air         & A.       & \ding{55} & 83 &  8,446 (T.\&E.) & S. & RGB-D \\
R2RIE-CE           & I.     & Residence   & A.       & \ding{51} & 19 & 11,764 (E.)     & S. & RGB-D \\
\midrule
A2A                & I.\&O. & Agriculture & G.\&A.   & \ding{55} & 46 &  1,560 (E.)     & S. & RGB \\
\rowcolor{gray!15}
A2A-MI (Ours)      & I.\&O. & Agriculture & G.\&A.   & \ding{51} & 46 &  4,680 (E.)     & S. & RGB \\
\bottomrule
\end{tabular}
}
\end{table*}

\begin{figure}[t]
  \vspace{10pt}
  \begin{subfigure}[b]{0.32\linewidth}
    \includegraphics[width=\linewidth]{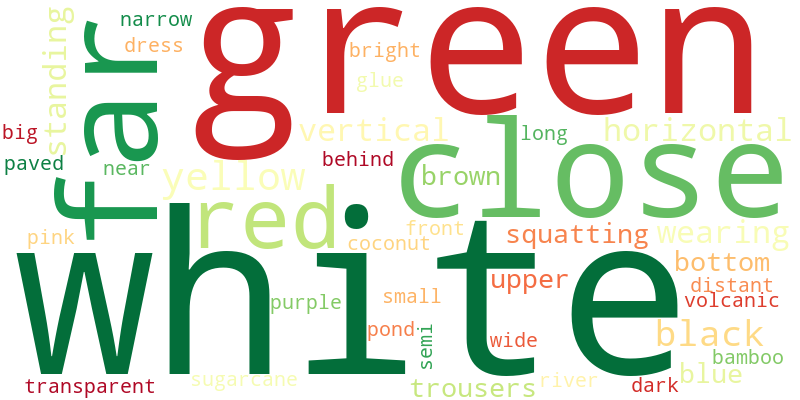}
    \caption{Adjectives.}
  \end{subfigure}
  \hfill
  \begin{subfigure}[b]{0.32\linewidth}
    \includegraphics[width=\linewidth]{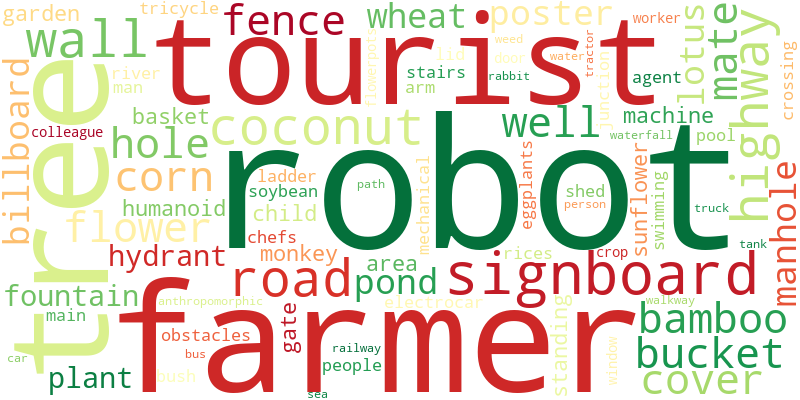}
    \caption{Nouns.}
  \end{subfigure}
  \hfill
  \begin{subfigure}[b]{0.32\linewidth}
    \includegraphics[width=\linewidth]{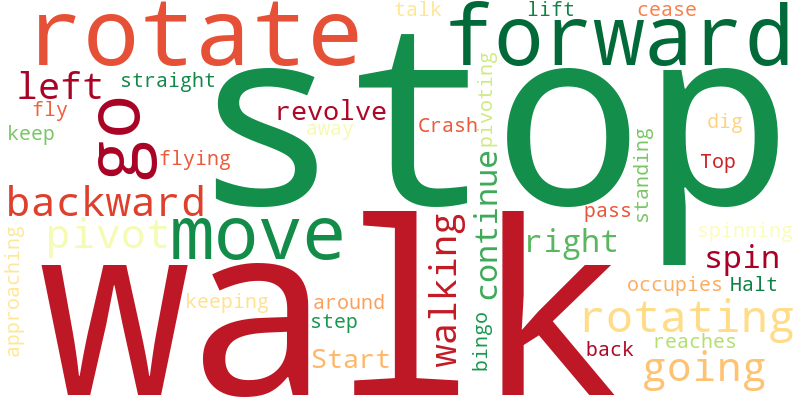}
    \caption{Verbs.}
  \end{subfigure}
  \caption{A2A-MI mistaken word distribution illustrations on different instruction mistake classes: The larger the font size, the higher the appearance frequency.
  }
  \label{fig:distribution}
\end{figure}

\subsection{Benchmark Assessment}
\label{benchmark_assessment}

\par We assess our A2A-MI benchmark with several mainstream VLN benchmarks \cite{CVPR:TOUCHDOWN,CVPR:R2R,CVPR:REVERIE,ECCV:VLN-CE,ICCV:AerialVLN,IROS:Error,arXiv:AgriVLN}, as shown in Table \ref{tab:benchmark_summary}. In terms of scenes and cameras, A2A-MI completely follow the A2A benchmark. In terms of instructions, each episode is respectively inserted with three classes of instruction mistakes, therefore, the final scale of A2A-MI is tripled to 4,680 instructions with an average length of 46 words. 
We further summarize the mistaken word distribution of A2A-MI, as illustrated in Fig. \ref{fig:distribution}. Whether on the adjective, none or verb portion, the new inserted mistaken words are practical and diversified, fully aligning with the potential speaking mistakes from the agricultural workers in the reality. 
In summary, we suggest A2A-MI as a large-scale and high-quality benchmark for effectively evaluating whether an agricultural VLN agent can be aware of instruction mistakes.

\section{Methodology}

\begin{figure*}[t]
\centering
\includegraphics[width=1.0\linewidth]{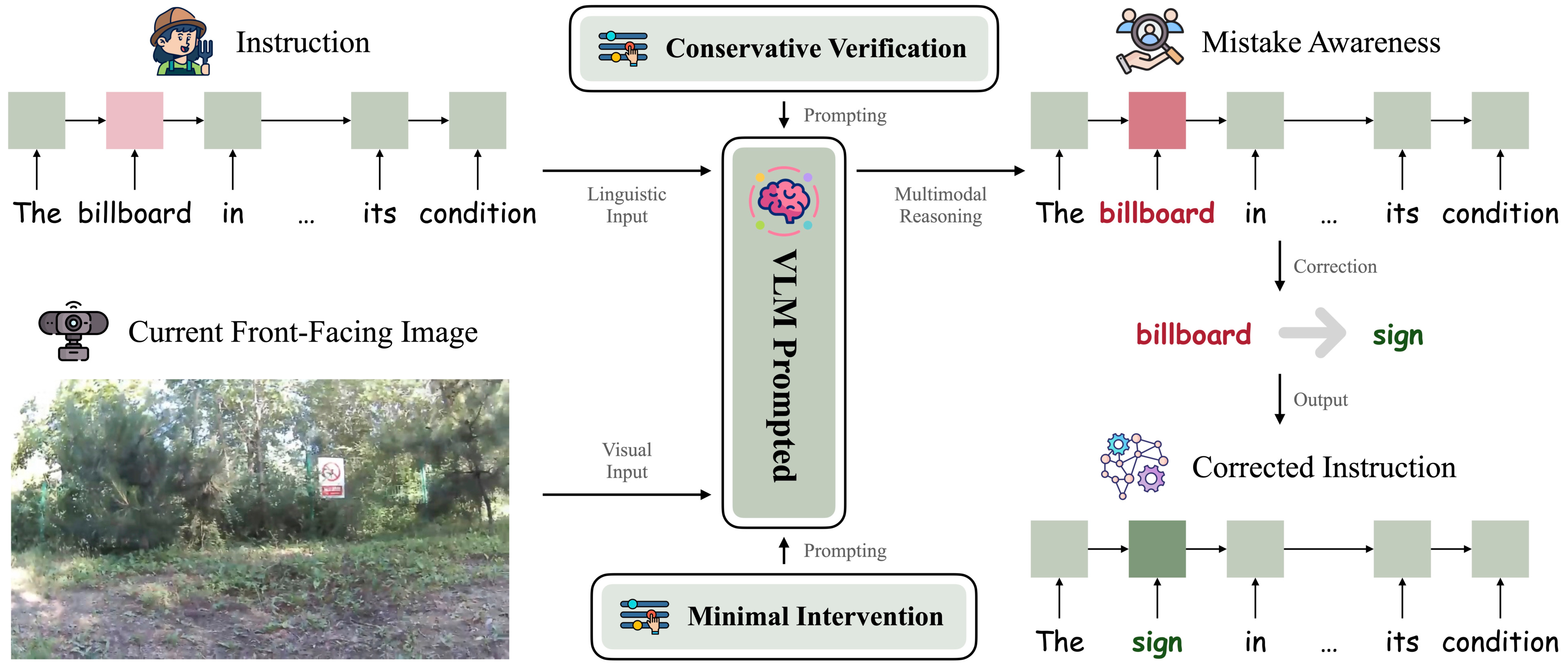}
\caption{IMAC-AgriVLN illustration: In our IMAC module, we design two core principles to prompt a VLM analyzing the instruction and image, to reason whether the instruction has mistakes and attempt to correct them when needed.
}
\label{fig:method}
\end{figure*}

In this section, we present the 
IMAC-AgriVLN method, as illustrated in Fig. \ref{fig:method}. First, we present the IMAC module in Sec. \ref{sec:IMAC}. Second, we integrate it into the AgriVLN backbone to build our IMAC-AgriVLN method in Sec. \ref{sec:base_model}.

\subsection{IMAC}
\label{sec:IMAC}

\par The inputs of IMAC are an instruction $W$ and an image $I$. 
We prompt a VLM denoted as $\text{IMAC}\,(\,\cdot\,)$ in a zero-shot manner, to reason whether the instruction has mistakes, defined as:
\begin{equation}
\label{eq:IMAC}
\hat{S}, \hat{W}, \rho 
= 
\text{IMAC} \big(P_{sys}^{\text{IMAC}}, P_{usr}^{\text{IMAC}} (W, I)\big)
\end{equation}
where $\hat{S}$ is the contiguous span with mistakes. $\hat{W}$ is the corrected instruction. $\rho$ is the decision summary, which provides an explicit interpretation. $P_{usr}^{\text{IMAC}}\,(\,\cdot\,)$ is the user prompt template for $\text{IMAC}\,(\,\cdot\,)$. $P_{sys}^{\text{IMAC}}$ is the system prompt template for $\text{IMAC}\,(\,\cdot\,)$, in which we design two core principles as follows: 
\begin{quote}
\textit{\textbf{Conservative Verification}: 
Only identify a mistake if you are highly confident that the instruction is incorrect and would likely cause navigation failure or unsafe behavior, based on the instruction and the current camera observation, which reflects only the robot's present viewpoint and may not include all relevant landmarks.
\\
\textbf{Minimal Intervention}: 
If an error exists, identify the smallest incorrect word or span, and correct it with the minimal necessary change only. Do not perform stylistic improvements or rephrasing.
}
\end{quote}
If $\hat{S} =$ “None”, i.e., no mistake is identified, $\text{IMAC}\,(\,\cdot\,)$ keeps silent. If $\hat{S} \neq$ “None”, i.e., the most possible mistake is identified, $\text{IMAC}\,(\,\cdot\,)$ attempts to correct the instruction from $W$ to $\hat{W}$ by replacing $\hat{S}$ with a contiguous span.

\subsection{Backbone}
\label{sec:base_model}

\par To build our IMAC-AgriVLN method, we follow AgriVLN \cite{arXiv:AgriVLN} as our backbone, and we integrate our IMAC module into two places. 

\par After receiving an instruction $W$, to begin with, we call IMAC for the first time to build the awareness on instruction mistakes, defined in Equation \ref{eq:IMAC}, in which we set $W$ and $I_0$ as the inputs, because a robot standing on the starting point only access the initial camera image. Then we keep using the original STL module denoted as $\mathcal{S}\,(\,\cdot\,)$, which decomposes the processed instruction $\hat{W}$ into a list of structured subtasks, defined as:
\begin{equation}
\left\{ (D_i, SC_i, EC_i, \sigma_i) \right\}_{i=1}^{N}
=
\mathcal{S}(\hat{W})
\end{equation}
where $D_i$, $SC_i$, $EC_i$ and $\sigma_i$ are the subtask description, starting condition, ending condition and current state of the $i$-th subtask, respectively. $N$ is the length of the subtask list.

\par Next, IMAC-AgriVLN begins navigating the robot to move following the subtask list but not the whole instruction. At each time step $t$, to begin with, we call IMAC for the second time to build the awareness on subtask mistakes, defined in Equation \ref{eq:IMAC}, in which we set $D_{i^*}$ and $I_t$ as the inputs. If a mistake is identified, we use the outputs to update $D_{i^*}$, $SC_{i^*}$ and $EC_{i^*}$, where $i^*$ is the order of the current executing subtask. Then we keep using the original decision-maker denoted as $\mathcal{D}\,(\,\cdot\,)$, which understands the current executing subtask and the current camera image to output the most appropriate low-level action $\hat{a_t}$, defined as:
\begin{equation}
\hat{a_t}, \Delta \sigma_{i^*, t}, \rho_t = \mathcal{D}(D_{i^*}, SC_{i^*}, EC_{i^*}, I_t)
\end{equation}
where $\Delta \sigma_{i^*, t}$ is the state transition signal. If $\mathcal{D}\,(\,\cdot\,)$ thinks the next unexecuted subtask should be started, it changes $\sigma_{i^*}$ from \texttt{[pending]} to \texttt{[doing]}. If $\mathcal{D}\,(\,\cdot\,)$ thinks the current executing subtask should be ended, it changes $\sigma_{i^*}$ from \texttt{[doing]} to \texttt{[done]}. $\rho_t$ is the decision summary, which provides an explicit interpretation.

\par As the time step $t$ goes by, the predicted action sequence $\langle \hat{a_{0}}, \hat{a_{1}}, \hat{a_{2}}, \dots, \hat{a_{t}} \rangle$ navigates the robot from the starting point to the target position. In an episode, $\mathcal{D}\,(\,\cdot\,)$ ends when one of the following conditions happens: 
\begin{quote}
1) $\hat{a_{t'}}$ = $\texttt{[STOP]}$; \\
2) The predicted action sequence $\langle \hat{a_{t'-\tau}}, \hat{a_{t'-\tau+1}}, \dots, \hat{a_{t'}} \rangle$ is deviated to the ground-truth action sequence $\langle a_{t'-\tau}, a_{t'-\tau+1}, \dots, a_{t'} \rangle$; \\
3) $t'$ reaches the max allowable limitation of time step.
\end{quote}
where $t'$ is the stopping time step. $\tau$ is the deviation time threshold, which is set to 4 s following the setting of the A2A \cite{arXiv:AgriVLN} benchmark.

\section{Experiments}

\subsection{Experimental Settings}
\label{sec:experimental_settings}
We implement all the experiments on our proposed A2A-MI benchmark. We locally deploy DeepSeek-R1-32B \cite{Nature:DeepSeek} with Q4\_K\_M quantization as the LLM and Qwen2.5-VL-32B \cite{arXiv:Qwen} with Q4\_K\_M quantization as the VLM. Regarding the hyper-parameters, we set the inference temperatures of the LLMs and VLMs to 1e-4 to ensure the response stability. All the experiments run on a single NVIDIA L20 GPU with 48G video memory.

\subsection{Evaluation Metrics}
\label{sec:evaluation_metrics}
We follow the two standard VLN evaluation metrics \cite{CVPR:R2R}: Success Rate (SR) and Navigation Error (NE). NE measures the path length between the stopping position and the target position. SR measures the rate successfully reaching the target position within a 2-m NE. Besides, we propose a new evaluation metric to specifically evaluate the effectiveness of our IMAC module, titled Instruction Mistake Valid Awareness Rate (AR). In a single episode, AR is calculated as:
\begin{equation}
AR =
\begin{cases}
1, & \text{if } \exists t \in \{0,\dots,t'\} \text{ s.t. } \hat{S}_t = \tilde{S}, \\
0, & \text{otherwise}.
\end{cases}
\end{equation}
where $\{ \hat{S}_t \}_{t=0}^{t'}$ is the set of predicted mistaken spans, and $\tilde{S}$ is the labeled mistaken span. We calculate the average score across all the episodes, defined as:
\begin{equation}
\overline{AR} = \frac{1}{Q} \sum_{e=1}^{Q} AR_e
\end{equation}
where $AR_e$ is the score of the $e$-th episode, and $Q$ is the episode quantity.

\subsection{Comparison Experiment}
\label{sec:comparison_experiment}

\par We evaluate our IMAC-AgriVLN method and three state-of-the-art methods \cite{EMNLP:SIA-VLN,RAL:DILLM-VLN,arXiv:AgriVLN} on our A2A-MI benchmark. Besides, the methods of Random, Fixed and Human are reproduced as the lower and upper bounds, respectively. 

\par The comparison results are shown in Table \ref{tab:comparison_experiment}. When any class of instruction mistake is inserted, we observe significant drops across all the methods, such as the AgriVLN baseline (\#5) decreases SR by 57\% in average and increases NE by 9\% in average, from which we suggest an answer to the opening question: \textbf{\textit{Agricultural VLN agents tend to assume that the given instructions are correct, so when the scenes they see do not align with the instructions they receive, they do not have the awareness to doubt the instructions, unless the instruction mistakes are extremely obvious.}}

\begin{table*}[t]
\caption{Comparison results between our IMAC-AgriVLN and state-of-the-art methods: 
\textbf{Bold} and \underline{underline} mark the best and worst scores, respectively.}
\label{tab:comparison_experiment}
\begin{center}
\renewcommand{\arraystretch}{1.2}
\setlength{\tabcolsep}{3pt}
\begin{tabular}{rlccccc|ccccc}
\toprule
\multirow{2}{*}{\textbf{\#}} & \multirow{2}{*}{\textbf{Method}} & 
\multicolumn{5}{c}{\textbf{SR} $\uparrow$} & 
\multicolumn{5}{c}{\textbf{NE} $\downarrow$} \\ 
\cmidrule(lr){3-7} \cmidrule(lr){8-12}
& & \ding{55} & \textbf{adj} & \textbf{nun} & \textbf{vrb} & \textbf{avg} &
\ding{55} & \textbf{adj} & \textbf{nun} & \textbf{vrb} & \textbf{avg} \\
\midrule
1 & Random       & 0.06 & 0.06 & 0.06 & 0.06 & 0.06 & 7.03 & 7.03 & 7.03 & 7.03 & 7.03 \\
2 & Fixed        & 0.03 & 0.03 & 0.03 & 0.03 & 0.03 & 3.06 & 3.06 & 3.06 & 3.06 & 3.06 \\
\midrule 
3 & SIA-VLN      & \underline{0.15} & \underline{0.09} & \underline{0.10} & \underline{0.07} & \underline{0.09} & \underline{4.52} & \underline{4.95} & \underline{4.94} & \underline{5.13} & \underline{5.01} \\ 
4 & DILLM-VLN    & 0.21             & 0.11             & \underline{0.10} & \underline{0.07} & \underline{0.09} & \textbf{3.99}    & \textbf{4.24}    & \textbf{4.17}    & \textbf{4.52}    & \textbf{4.31} \\ 
5 & AgriVLN      & \textbf{0.23}    & 0.10             & \underline{0.10} & 0.11             & 0.10             & 4.43             & 4.83             & 4.73             & 4.86             & 4.81 \\ 
\rowcolor{gray!15}
6 & IMAC-AgriVLN & 0.21             & \textbf{0.14}    & \textbf{0.15}    & \textbf{0.13}    & \textbf{0.14}    & 4.45             & 4.77             & 4.72             & 4.87             & 4.79 \\
\midrule
7 & Human        & 0.87 & 0.79 & 0.80 & 0.71 & 0.77 & 0.57 & 0.93 & 1.04 & 1.31 & 1.09 \\
\bottomrule
\end{tabular}
\end{center}
\vspace{-15pt}
\end{table*}

\begin{table}[t]
\caption{Ablation results on 
instruction mistake classes.}
\label{tab:ar}
\begin{center}
\renewcommand{\arraystretch}{1.2}
\setlength{\tabcolsep}{5pt}
\begin{tabular}{rlccccc}
\toprule
\multirow{2}{*}{\textbf{\#}} & \multirow{2}{*}{\textbf{Method}} & 
\multicolumn{5}{c}{\textbf{AR} $\uparrow$} \\
\cmidrule(lr){3-7}
& & \ding{55} & \textbf{adj} & \textbf{nun} & \textbf{vrb} & \textbf{avg} \\ 
\midrule
8 & IMAC-AgriVLN & - & 0.09 & 0.57 & 0.15 & 0.27 \\
\bottomrule
\end{tabular}
\end{center}
\vspace{-5pt} 
\end{table}

\begin{figure}[!t]
  \begin{subfigure}[b]{0.49\linewidth}
    \includegraphics[width=\linewidth]{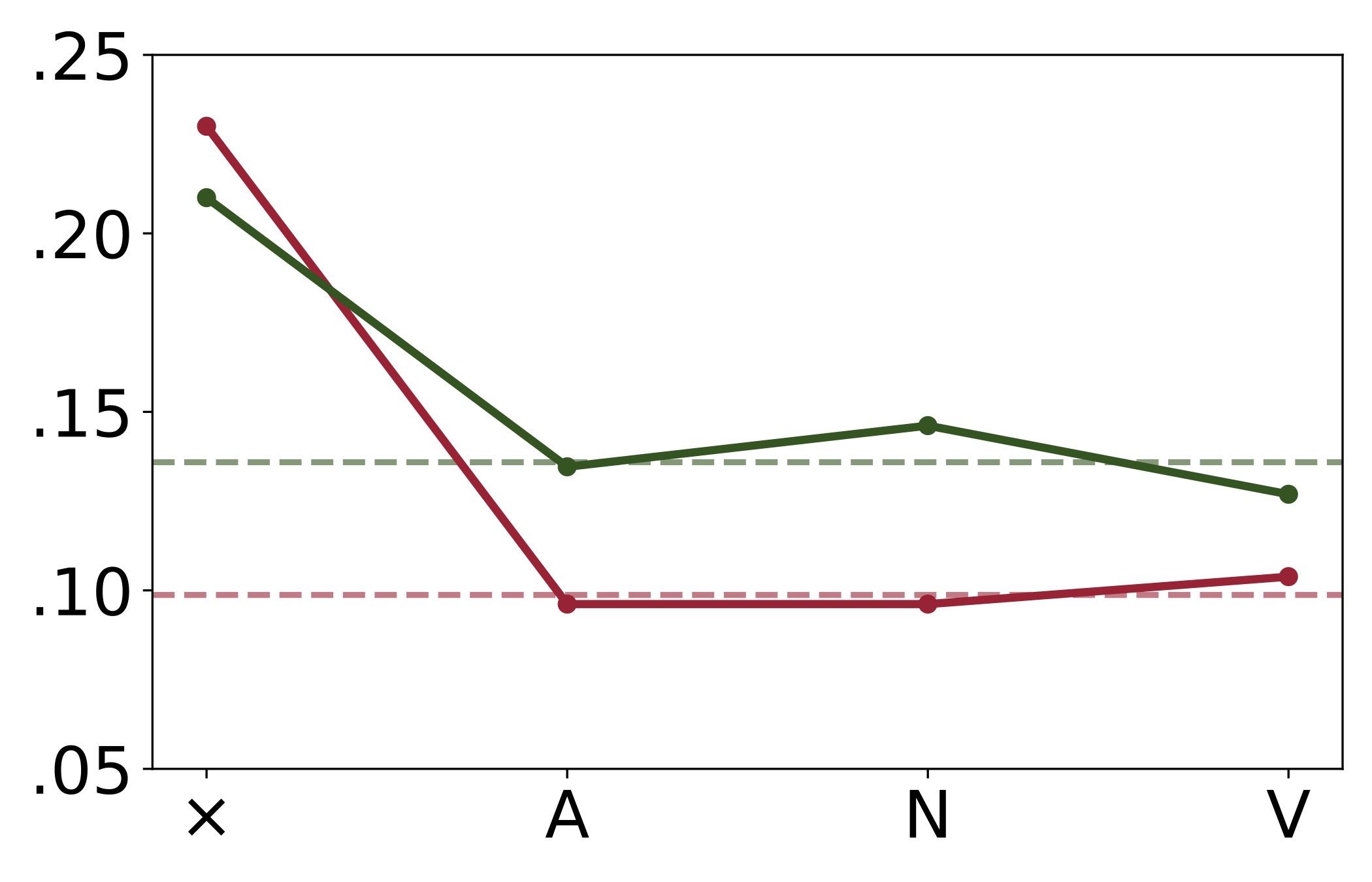}
    \caption{Success Rate $\uparrow$}
  \end{subfigure}
  \hfill
  \begin{subfigure}[b]{0.49\linewidth}
    \includegraphics[width=\linewidth]{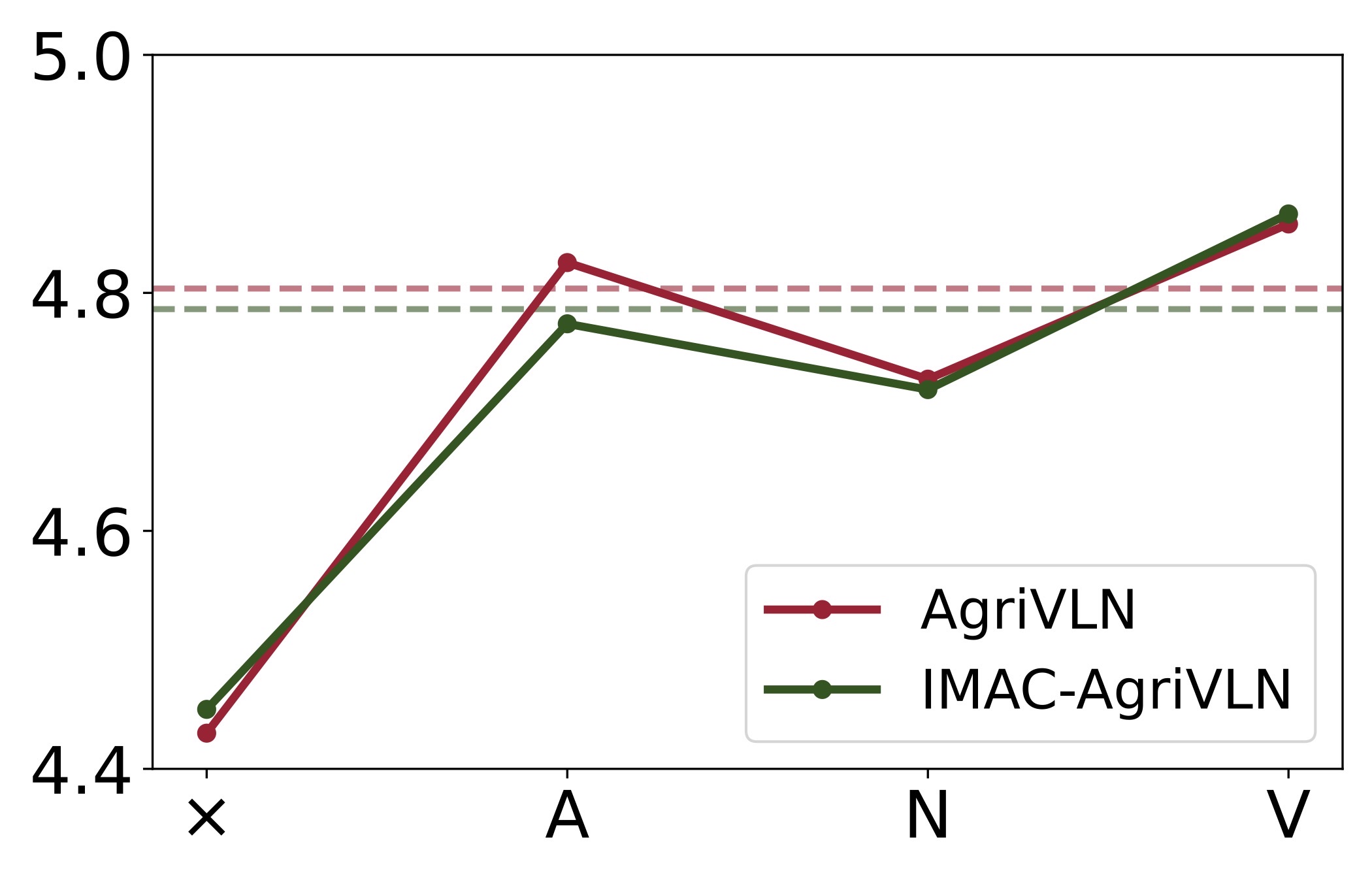}
    \caption{Navigation Error $\downarrow$}
  \end{subfigure}
  \caption {Ablation illustrations on our IMAC module.}
  \label{fig:comparison}
\end{figure}

\par As shown in Table \ref{tab:ar}, when our IMAC module is integrated, IMAC-AgriVLN achieves an average AR of 0.27, which indicates that more than a quarter of the instruction mistakes can be recognized and corrected by IMAC in a reasonable manner. In this way, SR is saved from 0.10 to 0.14 and NE is saved from 4.81 m to 4.79 m, which indicates that IMAC cannot perfectly restore a mistaken instruction, because there is still a gap to the performance when no mistake exists in the instructions. Nevertheless, this gap is sufficiently narrowing. Hence, we suggest that IMAC can correct instruction mistakes in an imperfect but already reasonable manner, which effectively assists the decision-maker to better understand human's true intentions from its mistaken instructions.

\begin{table}[t]
\caption{Ablation results between the STL module and our IMAC module: All the evaluation metrics are calculated in average across three instruction mistake classes. \textbf{Bold} and \underline{underline} mark the best and worst scores, respectively.
}
\label{tab:ablation}
\begin{center}
\renewcommand{\arraystretch}{1.2}
\setlength{\tabcolsep}{5pt}
\begin{tabular}{rlcc|ccc}
\toprule
\# & \textbf{Method} & \textbf{STL} & \textbf{IMAC} & \textbf{SR} $\uparrow$ & \textbf{NE} $\downarrow$ & \textbf{AR} $\uparrow$ \\ 
\midrule
9  & IMAC-AgriVLN & \ding{55} & \ding{55} & \underline{0.07} & \underline{5.39} & - \\
10 & IMAC-AgriVLN & \ding{51} & \ding{55} & 0.10             & 4.81             & - \\
11 & IMAC-AgriVLN & \ding{55} & \ding{51} & 0.13             & 5.00             & \underline{0.26} \\
\rowcolor{gray!15}
12 & IMAC-AgriVLN & \ding{51} & \ding{51} & \textbf{0.14}    & \textbf{4.79}    & \textbf{0.27} \\
\bottomrule
\end{tabular}
\end{center}
\vspace{-10pt} 
\end{table}

\subsection{Ablation Study}
\label{sec:ablation_study}

\subsubsection{Instruction Mistake Class}
\par 
We implement the ablation study on instruction mistake classes, as shown in Table \ref{tab:comparison_experiment} and Fig. \ref{fig:comparison}. For the AgriVLN baseline (\#5), SRs basically keep consistent, while NEs on the classes of adjective, noun and verb are 4.83 m, 4.73 m and 4.86 m, respectively, which indicates that verb mistakes tend to be more challenging and noun mistakes tend to be relatively acceptable. We attribute this difference to the semantic weight: 
In an agricultural VLN task, the final required output is a low-level action, such as \texttt{[LEFT ROTATE]}, therefore, behavioral verbs tend to provide the most direct references, such as \textit{turn left}. Relative speaking, descriptive adjectives and concrete nouns tend to provide references more like in an assistant manner. Hence, in an instruction, behavior verbs' semantic weights tend to be higher than descriptive adjectives' or concrete nouns'. We suggest this preference as a double-edged sword: When a behavior verb is correct, it can provide the most direct guidance. When a behavior verb is mistaken, however, it can also provide the most misleading information, and it is difficult for the decision-maker to be aware of. 

\subsubsection{IMAC v.s. STL}
\par Considering that the STL module also has a linguistic preprocessing effect, we further implement the ablation study on the contribution between IMAC and STL against instruction mistakes, for which we respectively remove them from our IMAC-AgriVLN method, as shown in Table \ref{tab:ablation}. When only STL is retained (\#10), SR increases by 43\% and NE decreases by 11\%, respectively. When only IMAC is retained (\#11), SR increases by 86\% and NE decreases by 8\%, respectively. When both modules are retained (\#12), the performance reaches to the best. Hence, we suggest that STL and IMAC corporate in a compatible manner, and the main contribution to build the awareness and correction on instruction mistakes stems from IMAC.

\begin{figure*}[t]
\centering
\includegraphics[width=1.0\linewidth]{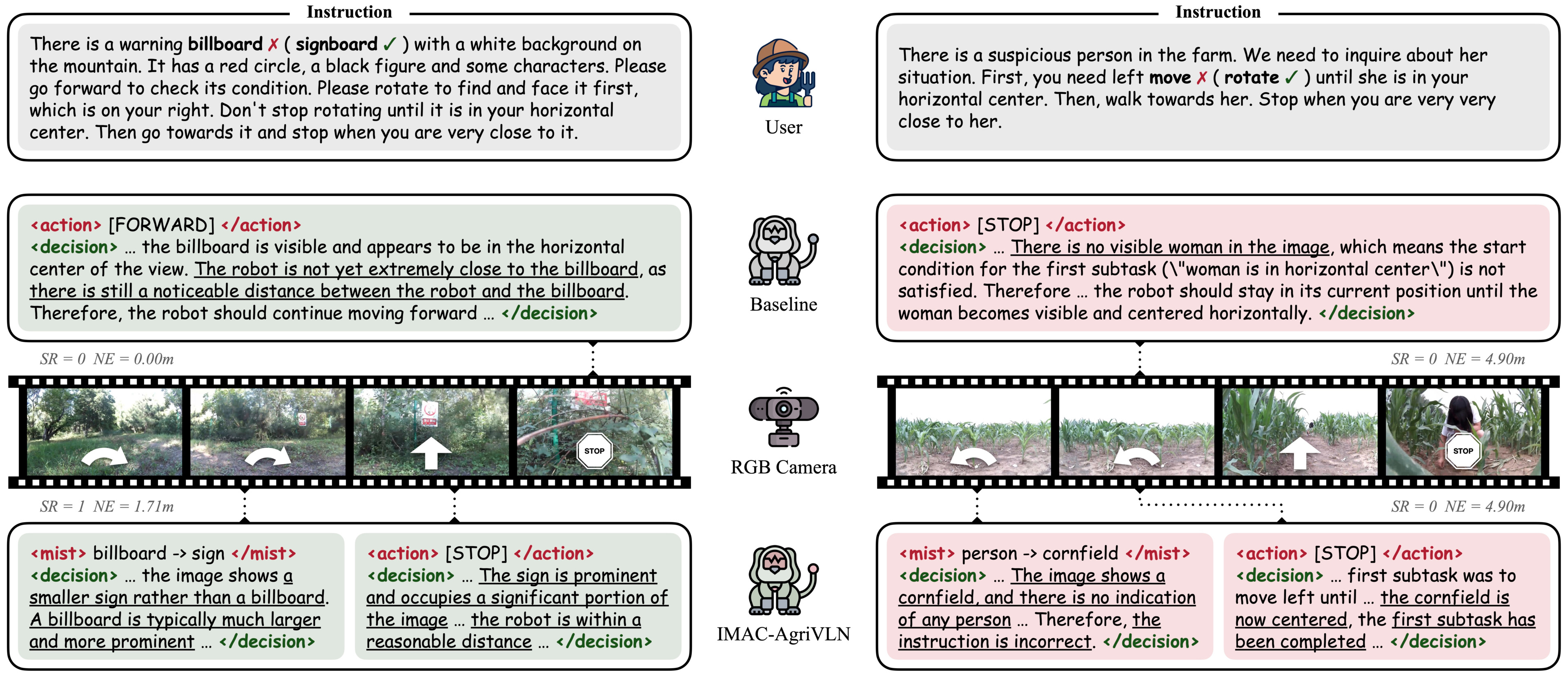}
\caption{Case study illustration on a successful case (left) and a failed case (right): In each case, from top to bottom are the mistaken instruction, the baseline's decision-making results, the RGB camera streaming with the ground-truth actions, and our IMAC-AgriVLN's decision-making results. In the mistaken instructions, \textcolor[HTML]{135416}{\ding{51}} and \textcolor[HTML]{B11E31}{\ding{55}} mark the original correct word and the revised mistaken word, respectively.}
\label{fig:qualitative}
\end{figure*}

\subsection{Case Study}
\label{sec:qualitative_experiment}

\par Towards a more comprehensive discussion, we share both a successful case and a failed case, as illustrated in Fig. \ref{fig:qualitative}. Please note that before the instruction mistakes are inserted, the AgriVLN baseline can succeed both cases. We mark all the decision-maker's pivotal reasoning thoughts with \textit{italics} in the text and \underline{underline} in the figure. 

\par In the successful case, the mistaken word in the instruction is a concrete noun, i.e., billboard \textcolor[HTML]{B11E31}{\ding{55}} (signboard \textcolor[HTML]{135416}{\ding{51}}). Until at the maximum time step t = 16.4 s, the baseline still thinks that \textit{“there is still a noticeable distance between the robot and the billboard”}, so \textit{“the robot is not yet extremely close to the billboard”}, thereby choosing to keep going \texttt{[FORWARD]}, which results in a crash to the fence (SR = 0, NE = 0.00 m). We attribute this crash to the incorrect recognition on object scale: Compared to a signboard, a billboard is usually much bigger. Since the billboard only occupies about half of the camera frame, the baseline's thinking is reasonable. In contrast, at the time step t = 5.4 s, the IMAC module thinks that \textit{“a billboard is typically much larger and more prominent”}, so builds the awareness that the target object should be \textit{“a smaller sign rather than a billboard”}, then corrects it in a reasonable manner, i.e., billboard \textcolor[HTML]{B11E31}{\ding{55}} $\rightarrow$ sign \textcolor[HTML]{135416}{\ding{51}}. At the time step t = 13.0 s, our IMAC-AgriVLN method observes that \textit{“the sign is prominent and occupies a significant portion of the image”}, so properly thinks that \textit{“the robot is within a reasonable distance”}, thereby predicting the correct \texttt{[STOP]} action (SR = 1, NE = 1.71 m, AR = 1). From this successful case, we suggest that a mistaken instruction may mislead the decision-maker on visual perceptions, and IMAC can build an effective awareness on instruction mistakes and correct them in a reasonable manner.

\par In the failed case, the mistaken word in the instruction is a behavior verb, i.e., move \textcolor[HTML]{B11E31}{\ding{55}} (rotate \textcolor[HTML]{135416}{\ding{51}}). At the time step t = 1.0 s, observing that \textit{“there is no visible woman in the image”}, the baseline should have chose \texttt{[LEFT ROTATE]} to find her. But the clear behavior verb “rotate” is replaced by an ambiguous word “move”, which makes it deviate from the correct path after the disorganized thinking (SR = 0, NE = 4.90 m). Unfortunately, IMAC also fails to be aware of it, even further misjudges the correct target noun “person” as a mistaken word. At the time step t = 0.8 s, IMAC-AgriVLN observes that \textit{“the image shows a cornfield and there is no indication of any person”}, then does not recognize that the person is located to the left, being outside the current camera view, but improperly judges that \textit{“the instruction is incorrect”}, which results in an unreasonable correction, i.e., person \textcolor[HTML]{135416}{\ding{51}} $\rightarrow$ cornfield \textcolor[HTML]{B11E31}{\ding{55}}. Consequently, at the next time step, it observes that \textit{“the cornfield is now centered”}, so thinks that \textit{“the first subtask has been completed”}, which results in a too early prediction to \texttt{[STOP]} (SR = 0, NE = 4.90 m, AR = 0). From this failed case, we suggest that overly aggressive instruction mistake awareness may introduce additional noises, which may mislead the subsequent decision-making and even exacerbate the navigation errors.

\section{Conclusion}

In this paper, we propose the A2A-MI benchmark, in which we follow A2A as the foundation benchmark and insert three classes of instruction mistakes. We use it to evaluate several state-of-the-art agricultural VLN agents, then observe sufficient drops across all of them, from which we suggest that standard agricultural VLN agents cannot be aware of instruction mistakes. To address this problem, we propose the IMAC module analyzing the instruction and image, to reason whether the instruction has mistakes and attempt to correct them when needed. We integrate it into the AgriVLN backbone to build our IMAC-AgriVLN method, successfully saving SR from 0.10 to 0.14 and NE from 4.81 m to 4.79 m, which demonstrates the effectiveness of IMAC on strengthening the robustness against instruction mistakes. 

\par During the experiments, we also find a main limitation of the lacking evaluation metric on instruction mistake correction quality, which will be discussed in the below Limitation section. 

\par In the future, in addition to the improvement on the existing limitation, we plan to explore whether the ambiguous traversability be one of the reasons to the gap between agricultural VLN agents and human, and propose a new traversability estimation alarm module to narrow this gap.

\section*{Limitation}

Our proposed IMAC module consists of two steps: awareness and correction. In Section \ref{sec:evaluation_metrics}, in addition to the standard metrics of SR and NE, we propose AR to evaluate the quality of mistake awareness. However, we have not yet thought of a suitable way to build an evaluation metric on the correction quality, because the concept of “appropriate correction” is abstract and not unique. For example, in our case study, one of the mistaken words is billboard \textcolor[HTML]{B11E31}{\ding{55}} (signboard \textcolor[HTML]{135416}{\ding{51}}), and IMAC correct billboard \textcolor[HTML]{B11E31}{\ding{55}} $\rightarrow$ sign \textcolor[HTML]{135416}{\ding{51}}, in which we judge that both sign and signboard belong to “appropriate correction”. Hence, a simple rule-based algorithm (such as for SR) is not suitable for evaluating the correction quality. A possible solution is building a heavyweight LLM with crafted prompt as the professional judge to evaluate the correction quality in a more reasonable and convincing manner.

\begin{credits}

\subsubsection{\ackname} 
This work is supported by the Sichuan Chengdu Modern Agricultural Industry Research Institute of China Agricultural University: Provincial and Municipal Agricultural Subsidy Funded Project; the Natural Science Foundation of Sichuan Province (2024NSFSC0389); and the Provincial and Municipal Agricultural Subsidy Special Funds for the Construction of CAU–SCCD Advanced Agricultural and Industrial Institute. This work is also supported by Xiaobei Zhao. Thanks to Chiang Mai, Chiang Rai, Bangkok, Chitwan, Lumbini and Pokhara for the impressive traveling experiences, giving us a chilled vibe for experiment and writing. Thanks to Yuanquan Xu, the inspiration to us.

\subsubsection{\discintname}
The authors have no competing interests to declare that are relevant to the content of this article. 

\end{credits}

\bibliographystyle{splncs04}
\bibliography{custom}

\end{document}